\pdfoutput=1

\documentclass[onecolumn]{cohere}

\title{Back to Basics: Revisiting REINFORCE Style Optimization for Learning from Human Feedback in LLMs}

\author{
    name={Arash Ahmadian},
    affiliation={Cohere For AI},
}
\author{
    name={Chris Cremer},
    affiliation={Cohere},
}
\author{
    name={Matthias Gallé},
    affiliation={Cohere},
}
\author{
    name={Marzieh Fadaee},
    affiliation={Cohere For AI},
}
\author{
    name={Julia Kreutzer},
    affiliation={Cohere For AI},
}
\author{
    name={Olivier Pietquin},
    affiliation={Cohere},
}\author{
    name={Ahmet Üstün},
    affiliation={Cohere For AI},
}
\author{
    name={Sara Hooker},
    affiliation={Cohere For AI},
}

\date{\today}

\usepackage{wrapfig}
\usepackage{arydshln}

\abstract{AI alignment in the shape of Reinforcement Learning from Human Feedback (RLHF) is increasingly treated as a crucial ingredient for high performance large language models. \textsc{Proximal Policy Optimization} (PPO) has been positioned by recent literature as the canonical method for the RL part of RLHF. However, it involves both high computational cost and sensitive hyperparameter tuning. We posit that most of the motivational principles that led to the development of PPO are less of a practical concern in RLHF and advocate for a less computationally expensive method that preserves and even increases performance. We revisit the \textit{formulation} of alignment from human preferences in the context of RL. Keeping simplicity as a guiding principle, we show that many components of PPO are unnecessary in an RLHF context and that far simpler REINFORCE-style optimization variants outperform both PPO and newly proposed ``RL-free'' methods such as DPO and RAFT. Our work suggests that careful adaptation to LLMs alignment characteristics enables benefiting from online RL optimization at low cost.
}

\begin{document}

\section{Introduction}

\begin{quote}
    \textit{I suppose it is tempting, if the only tool you
have is a hammer, to treat everything as if it
were a nail.} \textbf{--- Abraham Maslow, 1966.}
\end{quote}

State-of-art Large Language Models (LLMs) are typically pre-trained on tremendous amounts of text \citep{brown2020languageGPT3,openai2023GPT4,anil2023palm,touvron2023llama,touvron2023llama2,ustun2024aya} spanning trillions of tokens. These training corpora often contain many complex preferences, relations, and intentions that may not all be desirable for an LLM to exhibit. A question of great interest to both the research and wider practitioner community is \textit{how to align these models to human preferences?}

Despite being the focus of considerable research effort \citep{ouyang2022LLMRLHF,bai2022constitutional,lee2023rlaif,tunstall2023zephyr,khalifa2021distributional}, there is a lack of consensus regarding the optimal approach to achieve this goal. Reinforcement Learning from Human Feedback (RLHF), one of the most widely regarded alignment approaches, directly borrows from traditional RL literature and uses techniques such as \textsc{Proximal Policy Optimization (PPO)} to maximize the reward score produced by a reward model that is typically trained as a binary classifier on pairs of completions labeled by human annotators. While PPO has become a canonical approach cemented in popularity through its usage in the seminal literature on RLHF \citep{LearningToSummarizeHF,nakano2022webgpt,bai2022constitutional}, getting PPO to work in practice is non-trivial for non-RL specialists and comes with known issues:

\begin{enumerate}
  \item \textbf{Computational Cost:} $\:$ PPO typically requires loading up to 4 models simultaneously: the generator, the reference (for KL estimation), the critic, and the reward model, where the training of the generative and critic models are interleaved \citep{schulman2017proximal}. This challenge is further exacerbated by the size of modern LLMs, ranging in the billions of parameters \citep{openai2023GPT4,LearningToSummarizeHF,touvron2023llama}. 
\item \textbf{Optimization challenges:} $\:$ the unstable and sensitive nature of online RL optimization, and the relative algorithmic complexity of PPO requires niche expertise to tune it well \citep{engstrom2020implementation}.
\end{enumerate}

Recent works propose ``RL-free'' methods such as DPO \citep{rafailov2023DPO}, IPO \citep{azar2023IPO}  or iterative fine-tuning approaches to LLM preference training~\citep{yuan2023rrhf,zhao2023slichf,dong2023raft}. However, these works fail to question whether a simpler solution within an RL paradigm exists. Instead, all these approaches attempt to answer this question by stripping all RL components from RLHF and the difficulties that come with it \citep{rafailov2023DPO,zhao2023slichf,yuan2023rrhf,liu2023statistical,dong2023raft,azar2023IPO}. Iterative fine-tuning techniques rely solely on a powerful reward model to identify a subset of samples to train on, while DPO and IPO avoid both reinforcement learning and training a separate reward model by directly learning from human feedback. 

\textbf{In contrast to these approaches, we remain in the RL paradigm, but instead \textit{return to basics}}. The core question we seek to explore in this work is \textit{can we avoid the computational and optimization complexity of PPO while preserving performance?}. We isolate several key differences between traditional Deep-RL settings which originally motivated PPO and typical human-preference learning settings for LLMs. We note that PPO, as an approach, emphasizes stability across iterations, aiming to train an effective policy with the premise of \textit{small, stable updates}. PPO was designed for a regime where off-policy gradient updates are large enough to introduce instability. This regime dominates traditional Deep-RL benchmarks \citep{engstrom2020implementation,schulman2017proximal}. However, in this work, we posit that the setting of RLHF, which involves fine-tuning a pre-trained LLM, is lacking in these characteristics.

In contrast to traditional Deep-RL settings, the initialization of the policy, in the form of a pre-trained and supervised fine-tuned (SFT) model, is far from a random parameterization. While the conceivable search space is enormous, due to the pre-training and SFT stages, only a far smaller subset of tokens is likely to be generated as the probability mass is concentrated on these few tokens. Thus, while traditional Deep-RL settings require strong regularization to reduce the high variance of the gradient estimators; we observe empirically this is less of a practical concern in RLHF and motivate a less computationally expensive method that preserves robustness \citep{wu-etal-2018-study,kreutzer-etal-2021-offline}. 

\begin{figure*}
    \centering
    \includegraphics[width=0.45\textwidth]{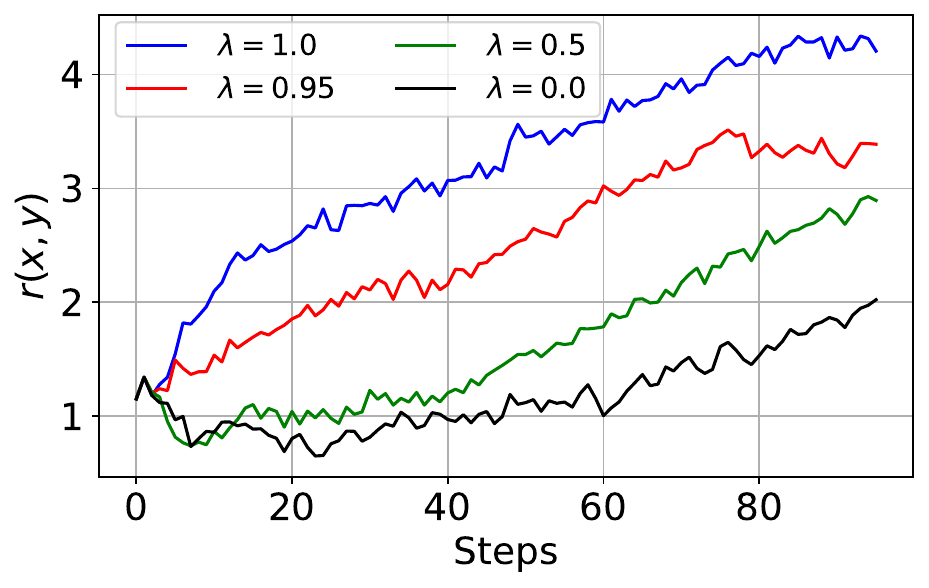}
    \includegraphics[width=0.45\textwidth]{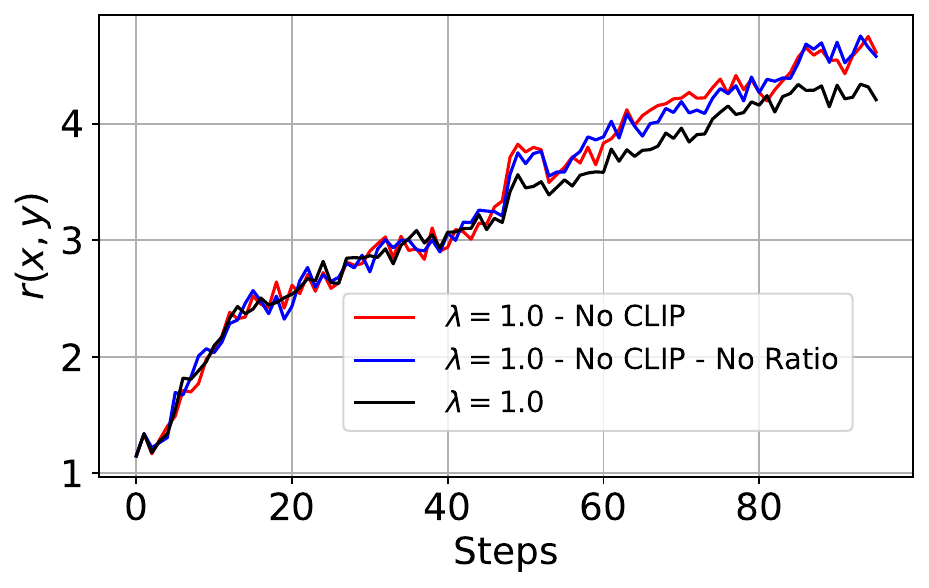}
    \caption{Training reward curves for PPO on Llama-7B based models with the HH dataset. \textbf{Left:} In an RLHF environment we observe that a higher emphasis on variance reduction at the cost of bias (low $\lambda$) performs worse than forgoing variance reduction but introducing less bias (high $\lambda$). $\lambda=1.0$ is the vanilla policy gradient which has the highest variance but no bias, and makes use of the full-trajectory reward in each update, performs the best. \textbf{Right:} PPO is unnecessarily complicated. Removing various augmentations of PPO such as clipping and loss normalization not degrade performance.}
    \label{fig:ppo_lambda}
\end{figure*}

Furthermore, we revisit how learning from human preferences is \textit{formulated} in the context of RL where generating each token is modeled as an \textit{action}, and each partial sequence, starting with the prompt, is seen as a \textit{state}. In practice, this modeling assumption for PPO method is often voided. 
We argue and show that the modeling of partial sequences is unnecessary in this setting where rewards are only attributed to \textit{full generations}, with no true rewards for any intermediary tokens in the generation. Thus, it is more appropriate and efficient to model the entire generation as a single action with the initial state determined by the prompt.

Given these observations, while keeping simplicity as a guiding principle, we explore the use of the \textsc{REINFORCE} estimator \citep{williams1992simple} and its multi-sample extension \textsc{REINFORCE Leave-One-Out (RLOO)} \citep{Kool2019Buy4R} to optimize the sequence-level objective. We break apart PPO and show that the most \textit{basic} policy gradient algorithm, Vanilla Policy Gradient REINFORCE consistently outperforms PPO. PPO is unnecessarily complicated for a pre-trained LLM environment. Unlike PPO, we can use REINFORCE to directly optimize the full trajectory (sequence) return coupled with unbiased baselines, whereas \textit{actor-critic} algorithms~\citep{sutton2000policygrad}, such as PPO, bootstrap off intermediary state value-functions to reduce variance at the cost of introducing bias into the estimator.

We arrive at consistent results across models including Llama \citep{touvron2023llama}, Pythia \citep{biderman2023pythia} and datasets such as the Anthropic Helpful \& Harmless \citep{bai2022AnthropicHH} and TL;DR Summarize \citep{LearningToSummarizeHF}:
\begin{enumerate}  

    \item \textbf{PPO is not the right tool for doing RL in RLHF.} We break apart PPO and show that the most ``basic'' policy gradient algorithm, Vanilla Policy Gradient REINFORCE \citep{sutton2020reinforcement}, is consistently outperforming PPO by 3.2\% to 20.3\% interms of win-rate, across all dataset and base model pairing .
    
    \item \textbf{RLOO outperforms key baselines.} Built on top of REINFORCE, RLOO enables using multiple online samples, and we empirically show it consistently outperforms baselines such as PPO, DPO \citep{rafailov2023DPO} as well as RAFT \citep{dong2023raft} across all datasets and models. We show that RLOO makes better use of online samples than RAFT while presenting a higher robustness to noise and degree of KL penalty. 
    
    \item \textbf{Modeling partial completions is not necessary.} We effectively demonstrate that modeling partial sequences is an unnecessary undertaking for LLM preference training. Instead, modeling the full generations preserves performance while reducing complexity in the RL stage and significantly accelerating learning. 

    \item  \textbf{RLOO is relatively robust to noise and KL penalty sensitivity.} We also accompany our results with a multi-dimensional analysis concerning language fluency, diversity, and robustness to noise. We showcase RLOO robustness to noise and degree of KL penalty compared to RAFT.

\end{enumerate}

\section{Background}
\label{sec:background}
\label{sec:rlhf}
The original RLHF pipeline for LLMs proposed in \citet{ziegler2020finetuning} consists of three stages: 

\textbf{(1) SFT Stage:} A pre-trained LM is instruction-tuned using a dataset consisting of a given instruction prompt, and (typically) a human-written completion. The LM/policy is trained with a cross-entropy loss over the completion only. Often, the SFT model, denoted as $\pi^{\text{sft}}$ is used to initialize both the reward model and the RLHF policy. 

\textbf{2) Reward Model Stage $\:$} RLHF methods
leverage a reward model $r_\phi(x,y)$ trained using a dataset of preferences $\mathcal{D}=\{(x,y_+,y_-)\}_{i=1}^N$ where $y_+$ and $y_-$ denote the preferred and not-preferred completions for the prompt $x$. The reward model is trained as a binary classifier with the following loss: 
\begin{equation}
    \mathcal{L}_{RM} = - \log \sigma(\log (r_\phi(x,y_+) - r_\phi(x,y_-)) 
\end{equation}

where $\sigma$ denotes the logistic function.

\textbf{(3) RL Stage:} 
In this stage, the reward model is used to provide online feedback in the optimization of the policy with the following objective: 

\begin{align}
     & \max_{\pi_\theta} \mathbb{E}_{x \sim \mathcal{D}, y\sim \pi_\theta(.|x)} 
     [r_\phi(x,y) - \beta D_{\text{KL}} \pi_\theta(.|x) || \pi_{\text{ref}}(.|x)]
\end{align}
where $\beta$ is meant to control the distance from the initial policy, $\pi_{\text{ref}}$ during the optimization of $r_\theta(x,y)$ as proposed in \citep{stiennon2022learningRLHF}. The KL-penalty is crucial as penalty-free optimization of the reward model leads to degradation in the coherence of the model.  Optimizing this objective is equivalent to maximizing the following KL-shaped reward in expectation: 
\begin{equation}
    R(x,y) = r_\phi(x,y) - \beta \log \frac{ \pi_{\theta}(y|x)}{\pi_{\text{ref}}(y|x)}
    \label{eq:fixed-objective}
\end{equation}

While reinforcement learning approaches share the components above, techniques differ in the formulation of the reward. To understand these differences, we introduce PPO and distinct alternatives such as REINFORCE and REINFORCE Leave-One-Out in the following sections. 

\subsection{PPO}\label{sec:ppo}
When using PPO in the RL stage, the initial state is determined by the prompt, each generated token is modeled as an action, and partial sequences are seen as states, with a discount factor ($\gamma \in [0,1]$) of 1 used. In this framework, only generating the \texttt{<EOS>} token carries a reward as output by the reward model which is combined with KL penalty, while for all other tokens in the vocabulary, only the KL component is non-zero: 
\begin{equation}
    \begin{split}        
    R(x,y) & = \sum\nolimits_{t=1} ^ {T} R_t(x,y_t)  
    \label{eq:brokendown-obj}
    \end{split}
\end{equation}
where $y_t$ denotes the $t$-th token of $y$, $T$ the number of tokens in the trajectory, and $R_i$ the correspondingly shaped reward. 

In practice, the following token-level clipped objective is used in PPO:
\begin{align}\label{eq:ppo_loss}   
     \min \Bigl(f(y_t|s_t) \hat{A}_\lambda(y_t,s_t), 
      & \text{clip}^{1+\epsilon}_{1-\epsilon}(f(y_t|s_t)) \hat{A}_\lambda(y_t,s_t) \Bigl)
      \text{  with } f(y_t|s_t)=\frac{\pi_\theta(y_t|s_t)}{\pi_{\text{old}}(y_t|s_t)},
\end{align}
where $s_t=\{y_{<t},x\}$ represents the state i.e. context at generation step $t$ that is composed of the history of generated tokens $y_{<t}$ and the given prompt $x$, $\pi_{old}$ is an older policy (not the same as $\pi_{ref}$), and $\hat{A}(y_t,s_t)$ is the estimated advantage function for generating token (action) $y_t$, at partial completion (state) at token $t-1$ of the generation, and $\epsilon$ is the clipping ratio. The advantage function is estimated using Generalized Advantage Estimation (GAE) \citep{schulman2018GAE}. 

\subsection{REINFORCE} \label{sec:REINFORCE_ESTIMATOR}

Given that in LLM applications,  $r(x,y)$ is only obtained \textit{at the end} of the full sequence, it may be more appropriate to model the \textit{entire generation} as a single action, as opposed to each token.  Although it has not been explored in the context of LLM alignment, modeling the full completion as a single action, as in the \textit{bandit} formulation, allows using the REINFORCE estimator \citep{kreutzer-etal-2017-bandit,nguyen2017reinforcement, williams1992simple}. This allows for back-propagating through the discrete action (generation) space, and directly optimize the KL-shaped reward objective 
for the entire sequence. 
    \begin{equation}\label{eq:reinforce}
        \mathbb{E}_{x \sim \mathcal{D}, y\sim \pi_\theta(.|x)}[R(y,x)\nabla_\theta \log \pi_\theta(y|x)]
    \end{equation}

To improve learning, one can reduce the variance of the estimator in Eq. \ref{eq:reinforce}, while keeping it unbiased, by subtracting a \textit{baseline} $b$ that has high covariance with the stochastic gradient estimate of Eq. \ref{eq:reinforce}~\citep{williams1992simple,mnih2014neural}:
    \begin{equation}
   \mathbb{E}_{x \sim \mathcal{D}, y\sim \pi_\theta(.|x)}[(R(y,x)-b)\nabla_\theta \log \pi_\theta(y|x)]
    \end{equation}

With a strong parameter-free choice for the baseline being the moving average of all rewards throughout training \citep{williams1992simple}:
    \begin{equation}
        b_{\text{MA}} = \frac{1}{S}\sum\nolimits_{s}R(x^s,y^s)
        \label{eq:MAbaseline}
    \end{equation}
    
Where $S$ is the number of training steps, and $(x^s,y^s)$ is the prompt-completion pair at the step $s$.  
    
\subsection{REINFORCE Leave-One-Out (RLOO)} 

The baseline in Eq.~\ref{eq:MAbaseline} is simple to implement and computationally cheap.  However, it can be improved upon if we have access to multiple online samples, that can be used for further unbiased variance reduction:
(1) The rewards for each sample can serve all other samples as a baseline. 
(2) Policy updates can be done on an average of gradient estimates for each sample, resulting in a variance-reduced multi-sample Monte-Carlo (MC) estimate. This is the intuition behind the REINFORCE Leave-One-Out (RLOO) estimator, proposed by \citep{Kool2019Buy4R}: 
    \begin{equation*}
        \begin{split}
        \frac{1}{k}&\sum_{i=1}^k [R(y_{(i)},x) - \frac{1}{k-1}\sum_{j\ne{i}}R(y_{(j)},x)] \nabla \log \pi(y_{(i)}|x) \text{      for   } y_{(1)},...,y_{(k)} \overset{i.i.d}{\sim} \pi_\theta (.|x)\\        
        \end{split}
        \label{eq:RLOO}
    \end{equation*}

Where $k$ refers to the number of online samples generated, $\textsc{RLOO}_{k}$ considers each $y_{(i)}$ individually and uses the remaining $k-1$ samples to create an unbiased estimate of the expected return for the prompt, akin to a \emph{parameter-free} value-function, but estimated at each training step. This is a much more effective baseline (as our experiments will show) than $b_{MA}$ since it's created on-the-fly for each sample and at each training step, but comes at a cost of increased sampling time during training. We note that generating extra samples as a means of variance reduction has been proposed by concurrent work \citep{li2023remax}, but we focus on RLOO here because of the efficiency benefits of fully utilizing all samples. 

\subsection{Alternatives to RL in Preference Training}

In the context of RLHF a substantial number of works propose ``RL-free'' methods which do not involve stage 3. We will benchmark RL approaches such as PPO, REINFORCE, and RLOO against these alternative methods such as ``Direct Preference Optimization (DPO)'' and RAFT \citep{dong2023raft}.We briefly introduce both below.

\noindent\textbf{Iterative Fine-tuning $\:$}
\label{sec:ift}
Iterative fine-tuning methods use the trained reward model to rank completions of online or offline sampled prompts, and then iteratively fine-tune the policy on a selected subset \citep{gulcehre2023reinforced,dong2023raft}. We note that this trick of using rewards from reinforcement/bandit learning coupled with supervised learning objectives is also known as \textit{bandit-to-supervised conversion} and had empirical success in offline RL for NLP problems with large action spaces before RLHF with LLMs~\citep{lawrence-riezler-2018-improving,kreutzer-etal-2018-neural}. 

We benchmark Reward rAnked FineTuning \citep[RAFT;][]{dong2023raft}, a simple cross-entropy loss is used on the best-ranked completion out of $k$ online samples, based on  $R(x,y)$ or $r(x,y)$. We note that RAFT does not make full use of all samples because it only optimizes using filtered top-ranked samples. In contrast, RLOO fully leverages constructing a baseline and a multi-sample MC estimate for the policy gradient.

\textbf{Direct Preference Optimization (DPO)$\:$}
Unlike other methods, DPO \citep{rafailov2023DPO} skips the reward modeling stage in the traditional RLHF pipeline and uses preference pairs to directly optimize the policy with the following loss: 
\begin{equation*}
    - \log \sigma(\beta \log \frac{ \pi_\theta(y_+|x)}{\pi_{\text{ref}}(y_+|x)} - \beta \log \frac{ \pi_\theta(y_-|x)}{\pi_{\text{ref}}(y_-|x)}) 
\end{equation*}

\section{From PPO to REINFORCE}
We scrutinize individual components of PPO that we consider not an ideal fit for RLHF. We explain the theoretical origin, motivate with the practical conditions of LLM RLHF, and provide empirical support from preliminary experiments. 

\subsection{Revisiting the need for low-variance estimators}

Actor-critic algorithms, such as PPO, were motivated in formulation by the high variance observed in traditional RL settings. PPO leverages lower-variance estimators of the total trajectory return to improve learning. These estimators are constructed by bootstrapping off a state-value function~\citep{sutton2000policygrad,schulman2018GAE,sutton2020reinforcement}. While bootstrapping reduces variance, the trade-off is the introduction of \textit{bias} which risks optimizing for biased rewards. 

In contrast, \textsc{REINFORCE} uses an unbiased Monte-Carlo estimator of the trajectory return that can have high variance in theory, especially if it is approximated with only a single sample, which is not frequently preferred in traditional Deep-RL environments. Recent work has offered a plethora of evidence that \textsc{REINFORCE} suffers from high variance and fails in the presence of large action spaces like NLP~\citep{ranzato2015sequence,bahdanau2017actorcritic,ding2017cold,ammanabrolu2020graph,ammanabrolu-etal-2022-aligning,martin-etal-2022-learning,NEURIPS2022_67496dfa}. However, we note that these findings were based on scenarios with \textit{poor conditioning} when \textit{training from random or weak initialization} as opposed to warm-starting it from a strong pre-trained model.

\textbf{Here, we question whether this empirical evidence holds for RLHF.} We posit that this is not a \textit{practical concern} in fine-tuning LLMs due to the extremely strong initialization of the policy (a pre-trained LLM). In this setting, strong initialization coupled with prompt conditioning leads to the concentration of probability mass on a few tokens at each generation step, even though the number of possible actions is in theory enormous (refer to Appendix \ref{app:conditioning} for further discussion on the effect of conditioning). The optimization landscape is far less likely to present problems like destructively large and high-variance gradient updates. Thus, attempting to reduce variance further at the cost of introducing bias is not worth it.

\textbf{Empirical support $\:$} To validate this hypothesis, we vary the weight placed upon variance minimization and bias introduction. In the formulation of PPO in Sect.~\ref{sec:ppo}, the advantage estimator GAE \citep{schulman2018GAE} is relied upon to trade-off bias and variance when estimating the true advantage function in PPO \citep{schulman2018GAE}. 
 
GAE introduces a hyper-parameter $\lambda \in [0,1]$ in the true advantage function, which balances bias and variance of the constructed estimator. The closer $\lambda$ is to 1, the higher the observed variance. The optimal choice of where to set  $\lambda=0$ depends on the environment. In a highly stochastic environment, minimizing variance at the cost of bias is a worthy trade-off. However, given a stable environment where variance is already low, the introduction of bias is needless. 

At the extreme of 1 which imposes minimal bias at the trade-off of variance, the advantage term reduces to return the estimator used in Vanilla Policy Gradient (PG) REINFORCE, which directly builds on the REINFORCE estimator, by optimizing the trajectory returns starting from each token in the generation   
\begin{equation}
     \sum\nolimits_{i=t} ^ {T} \gamma^{T-i-1}R_t(x,y_t) - b_{\phi}(s_t)
\end{equation}

where $b_{\phi}(s_t)$ is a learned baseline state $s_t$, akin to how a value network is learned in the traditional RL setting, using a standard MLE loss $\frac{1}{2} (\sum_{i=t} ^ {T} \gamma^{T-i-1}R_i(x,y_i) - b_\psi(s_t))^2$. Note that the key distinguishing factor between Vanilla PG and REINFORCE as referred to in this work, is that Vanilla PG uses the REINFORCE estimator on the trajectory return starting from the context formed by the prompt and \textit{a partial completion}, whereas the REINFORCE estimator described in Section \ref{sec:REINFORCE_ESTIMATOR} is applied to the \textit{the full trajectory return}. We will return to this distinction in the results Section \ref{sec:reward_optimization} when we evaluate whether evaluating partial completions is necessary in RLHF.

\textbf{In Figure~\ref{fig:ppo_lambda}, we present the results of evaluating the reward for PPO given  GAE with different value of $\lambda$}. Two variants impose minimal bias but invite high variance (($\hat{A}_{\lambda=1.00}$ (Vanilla PG introduced above), and $\hat{A}_{\lambda=0.95}$) and two variants which over-index on minimizing variance at the cost of bias ($\hat{A}_{\lambda=0.0}$,and $\hat{A}_{\lambda=0.5}$).   Figure~\ref{fig:ppo_lambda} plots the reward and observe that the most extreme variant Vanilla PG (unbiased $A_{\lambda=1.0}$) performs the best given it presents no bias at the risk of high variance. We observe a monotonically decreasing reward with decreasing $\lambda$. This supports our hypothesis that reducing variance at the cost of bias in an RHLF setting needlessly introduces bias given the stable default properties of the environment.

 \subsection{Clipping is Rarely Necessary in RLHF}
 Next, we turn to the clipping-ratio $\epsilon$ (see Eq.~\ref{eq:ppo_loss}), which is used to prevent large policy updates when $\frac{\pi_\theta}{\pi_{old}}$ deviates far from 1, i.e., to prevent updates that are too far off from the current policy \citep{schulman2017proximal}. 

\textbf{In Figure \ref{fig:ppo_lambda}, we compare reward curves for independent PPO training with and without clipping}. Note that we also turn off clipping for the value network, for these set of experiments, as it has been observed to have a noticeable impact on learning in traditional Deep-RL environments \citep{engstrom2020implementation}. The removal of these components does not impact learning meaningfully. We empirically found in our RLHF setting that the  loss is actually clipped on average $<5$\% of the time per batch, throughout training  across all dataset and base-model pairings, which indicates that the learning regime is close to being ``on-policy'', with policies varying slowly from one iteration to the other.

To further validate this, we completely turn off clipping followed by removing the ratio $\frac{\pi_\theta}{\pi_{old}}$, while $\lambda=1$, which reduces the PPO loss to that of Vanilla PG. If anything the removal of clipping gives a slight boost in performance, validating our hypothesis that large off-policy updates in our optimization regime are rare and do not have catastrophic effects on learning as they do in traditional Deep-RL. 

\subsection{Modeling Partial Completions is Not Necessary}

As described in Sect. \ref{sec:background}, PPO models each token as an action whereas REINFORCE models the \textit{entire generation} as a single action, as opposed to each token. In practice, in LLM RLHF a $r(x,y)$ is only attributed to the \texttt{<EOS>} token, where for all other tokens, only $\log\frac{\pi(y_t|s_t)}{\pi_{\text{ref}}(y_t|s_t)}$ composes $R_t(x,y)$, which is not meaningful. 

From a pure RL point of view, the environment dynamics are fully deterministic ($P_{D}(\{y_{<t+1,x}\}|s_t,y_t)=1$), meaning that our environment (context) changes deterministically based on the new token/action predicted. Hence, the problem can be reduced to a \textit{bandit} problem, where the Markov Decision Process (MDP) consists of only the initial state as determined by the prompt, and the terminal state, which is always reached after the generation \citep{kreutzer-etal-2017-bandit,nguyen-etal-2017-reinforcement}. Note that modeling the entire generation as a single action is done explicitly by REINFORCE but is also done implicitly with \textit{iterative fine-tuning} methods which generate the entire completion before filtering using a reward model.

In the \hyperref[sec:result]{results section} \ref{sec:reward_optimization} we will explicitly compare REINFORCE and RLOO which both model the \textit{full trajectory return} to PPO and Vanilla PG which both model the \textit{partial completion}. We ask \textbf{is modeling the entire generation as a single action sufficient to achieve similar or better performance in RLHF?}
    
\section{Experimental Setup}
\subsection{Training Details}

\textbf{Datasets $\:$} We report results on the TL;DR Summarize \citep{LearningToSummarizeHF} and Anthropic Helpful and Harmless Dialogue \citep{bai2022AnthropicHH} datasets. The trainig split of TL;DR Summarize \footnote{\url{https://github.com/openai/summarize-from-feedback}} dataset contains 116k human-written instructions and 93k human-annotated preference pairs. The preprocessed Anthropic-HH\footnote{\url{https://huggingface.co/datasets/Dahoas/full-hh-rlhf}} dataset contains 112k training preference pairs.

\textbf{Models $\:$} For both datasets, we use Pythia-6.9B \citep{biderman2023pythia} as the pretrained base-model. To ablate the effect of the pre-trained model quality on learning from human preferences, we also experiment with Llama-7B \citep{touvron2023llama} coupled with the Anthropic-HH dataset. 
 
To ensure a fair comparison across all methods, we use a context length of 512 tokens during both supervised fine-tuning and the reward model training. We initialize both the reward model and policy with the corresponding SFT checkpoint, unless noted otherwise.  

\textbf{Experimental Details} For the TL;DR Summarize dataset, we use the dedicated SFT split. Since the original Antrophic-HH dataset does not include a separate SFT split, we use prompts and the preferred responses from the binary comparisons during the SFT stage similar to prior work \citep{yuan2023rrhf,dong2023raft,rafailov2023DPO}. In the preference training stage, we use the same prompts as in the SFT stage to generate completions. Further details on the experimental setup and hyper-parameters are given in Appendix \ref{app:hparams} 

\begin{figure*}[!ht]
    \centering
    \includegraphics[width=1.0\textwidth]{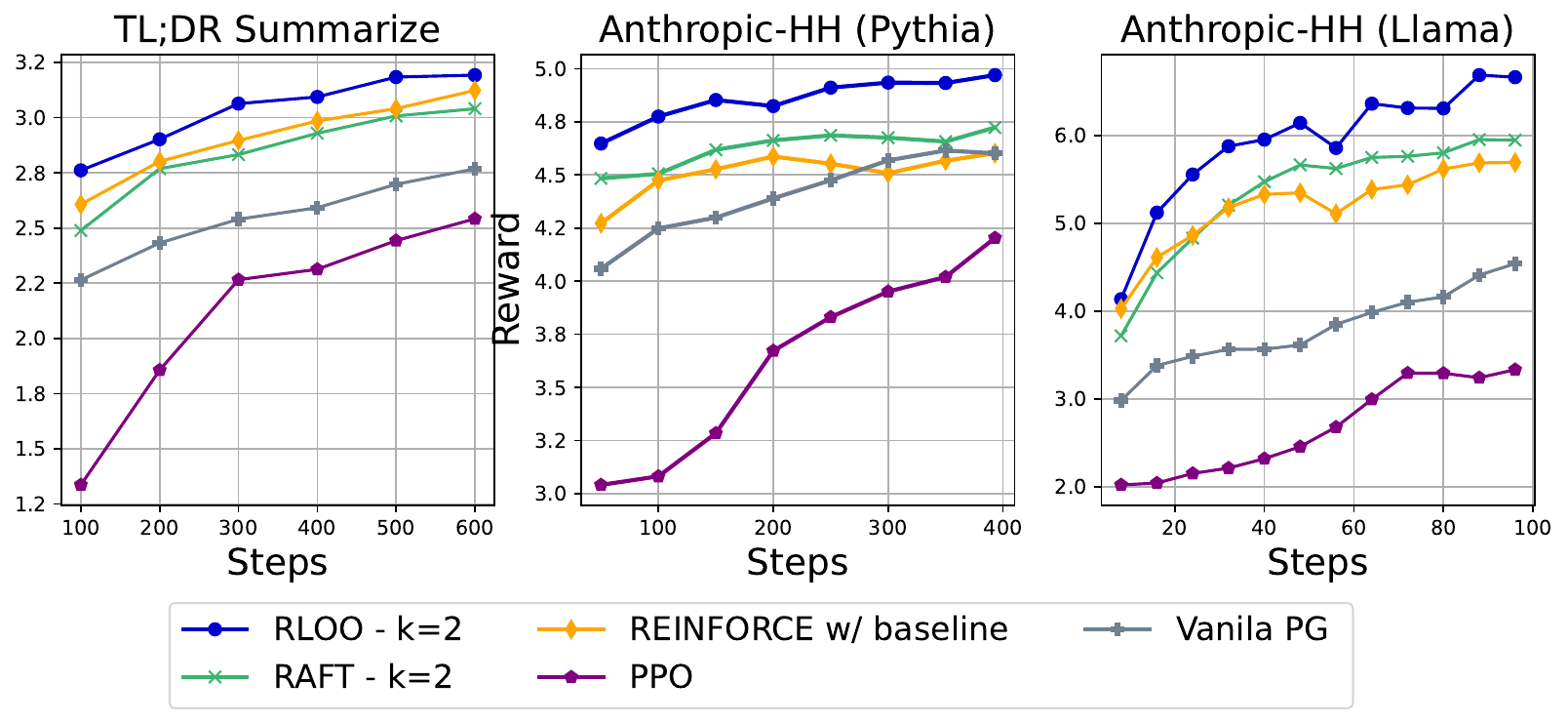}
    \caption{\textbf{Test rewards plotted throughout training.} RLOO outperforms all other methods consistently, while Vanila PG consistently outperforms PPO. REINFORCE w/ baseline refers to REINFORCE with a moving average reward baseline as in \hyperref[eq:MAbaseline]{Equation 8}}
    \label{fig:test_rewards_k12}
\end{figure*}

\begin{figure*}[!ht]
    \centering
    \includegraphics[width=1.0\textwidth]{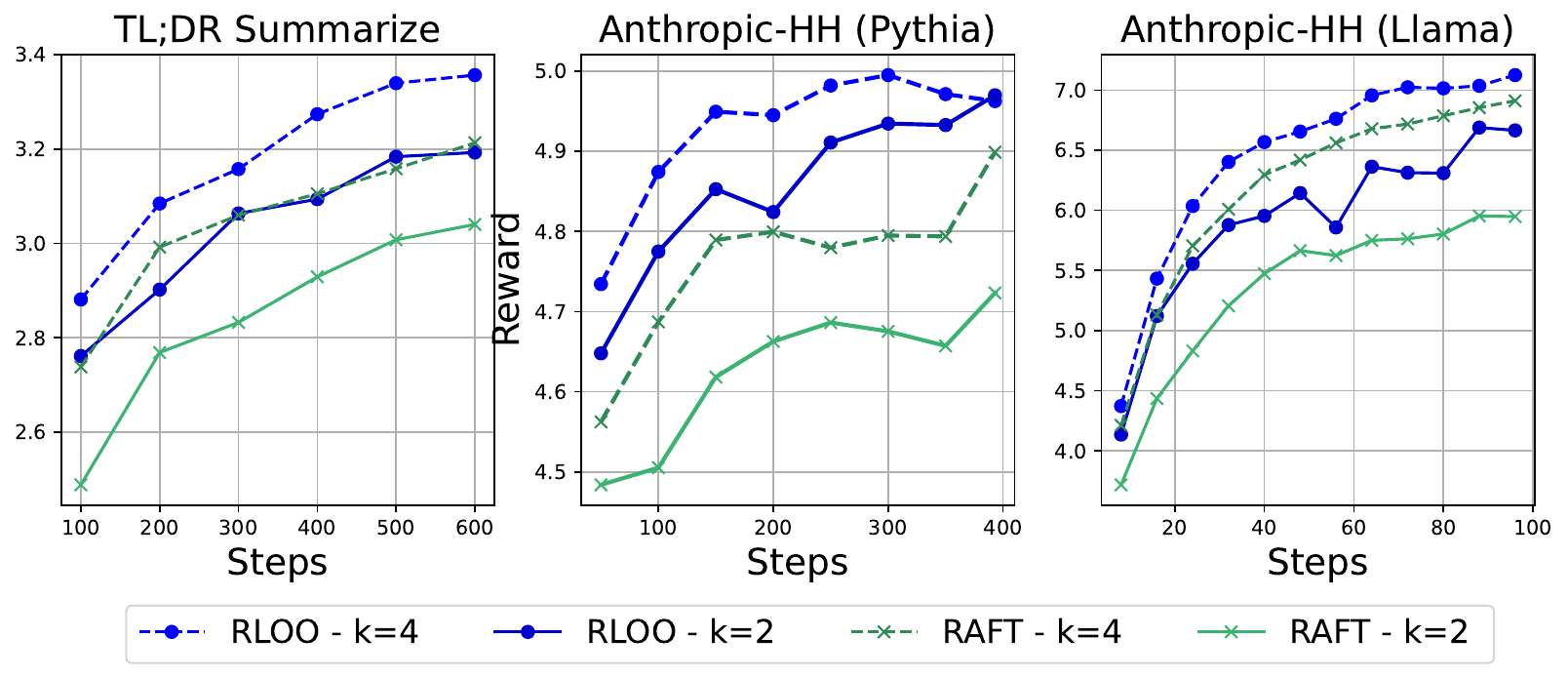}
    \caption{\textbf{Comparing Sample Efficiency} RLOO vs RAFT plot of test reward throughout training with $k=\{2,4\}$. For both values of $k$, RLOO outperforms RAFT given the same budget. RLOO$_{k=2}$ outperforms/matches RAFT$_{k=4}$ on both datasets, with a Pythia base model.}
    \label{fig:test_rewards_k4}
\end{figure*}

\subsection{Evaluation}

\textbf{Optimization Quality $\:$} For all online methods (all methods except DPO), to measure how well the method optimizes the \textit{intrinsic} objective , we report average rewards (using the training RM) on 1000 samples from the test set. To measure how well each method optimizes the \textit{extrinsic} objective of aligning the models to human preference, on the same test samples, we report simulated win-rates in accordance with Alpacafarm framework \citep{dubois2024alpacafarm} where we use GPT-4 as a proxy for human evaluation. We measure win-rates against reference SFT completions for the TL;DR dataset, and preferred completions for the HH dataset. At evaluation, we use greedy sampling unless otherwise noted. 

\textbf{Alignment Tax $\:$} RLHF fine-tuning is often associated with a drop in diversity and language fluency which is referred to as \textit{alignment tax} \citep{askell2021general,kirk2024understandingRLHF}. Hence, we also report metrics that serve as proxies for fluency and diversity similar to \citep{dong2023raft}.
To measure fluency, we report perplexity measured using the preferred completions from the test set, similar to \citet{dong2023raft}. Finally, we measure average completion length and diversity using average n-gram diversity \citep{li-etal-2016-diversity}.

\section{Results and Discussion}
\label{sec:result}

\subsection{Reward Optimization}\label{sec:reward_optimization} 

The objective of RLOO, REINFORCE with a (moving average) baseline, RAFT, PPO, and vanilla PG is to maximize the reward score, hence we compare the success of optimization for each method. On each dataset and base-model pair, we use the same reward model for all the methods, thus their test reward scores are directly comparable.

\textbf{Modeling Partial Completions vs Full Generations $\:$} As shown in Figure \ref{fig:test_rewards_k12}, we find that methods that do not model partial completions such as REINFORCE with baseline and RLOO, consistently outperform Vanilla PG and PPO where each token is modeled as an action (i.e. partial completions). Moreover, aside from their superior performance in reward optimization, these methods require loading one less model copy compared to Vanilla PG and PPO, and different means to create a baseline. This is because they eliminate the need for training a learned baseline and a value network, as required in Vanilla PG and PPO, respectively. This suggests that modeling partial sequences is unnecessary in the context of RLHF.

\textbf{Sampling Efficiency} Given the same sampling budget ($k$ online samples for each prompt), RLOO consistently outperforms RAFT throughout training as depicted in Figure \ref{fig:test_rewards_k4}. Noticeably, despite a smaller sampling budget, RLOO$_{k=2}$ either closely matches or outperforms RAFT$_{k=4}$, across all datasets and models. In this setting, RLOO uses only half the online-sample budget compared to RAFT given the same step count.      

\begin{figure}[t] 
    \centering
    \includegraphics[width=1.0\textwidth]{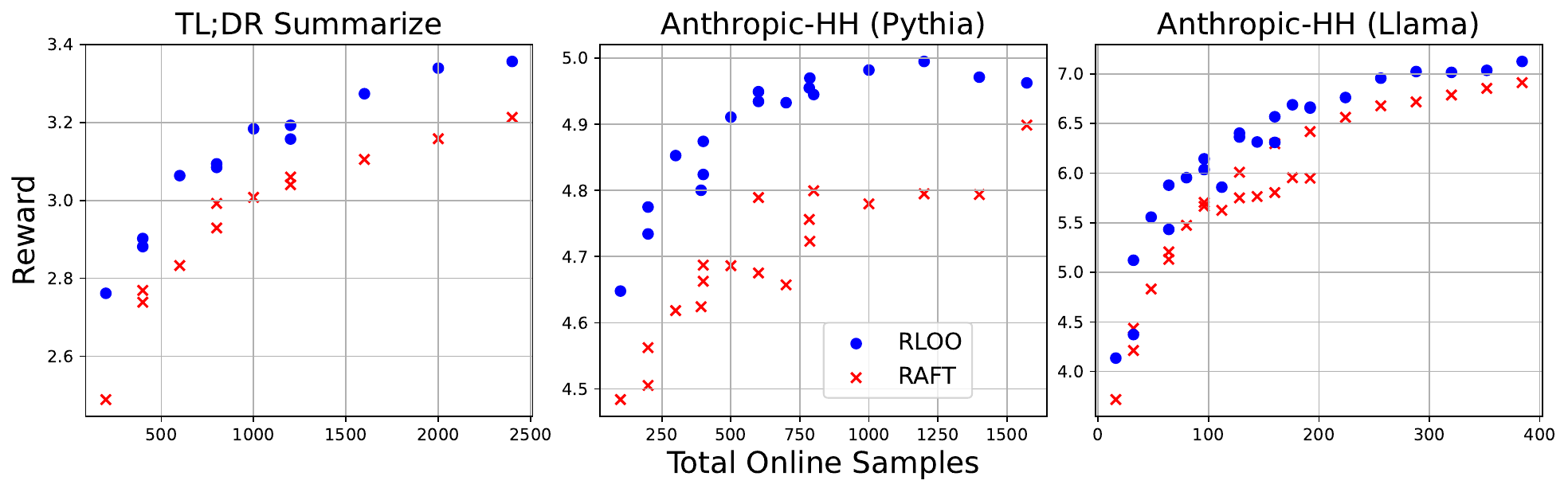}    

  \caption{ \textbf{Sample Efficiency} RLOO and RAFT test rewards are plotted at various points throughout training based on the total number of samples seen normalized by batch size (regardless of whether $k=2$ or $k=4$). RLOO is more efficient than RAFT in terms of online sample use, in both datasets and models.
  }
  \label{fig:RLOO_EFFICIENCY}
\end{figure}

This confirms that RLOO leads to better optimization by using all the samples generated, unlike RAFT where only the top-ranked sample is used for fine-tuning. Showcasing the same finding, Figure~\ref{fig:RLOO_EFFICIENCY} plots the rewards with respect to the number of samples generated during training regardless of the $k$ value.

\begin{table}[t]
    \centering
    \small
    \begin{tabular}{@{}lcccc@{}}
        \toprule
          Method  & TL;DR & HH (Pythia) & HH (Llama) \\ 
          \midrule
         \textsc{RLOO} (k=4) & \textbf{77.9}  & \textbf{43.7} & \textbf{64.1} \\
         \textsc{RAFT} (k=4) & 73.2  & 42.1 & 63.3 \\
         \noalign{\smallskip}
         \hdashline
        \noalign{\smallskip}
         \textsc{RLOO} (k=2) & \textbf{74.2}  & \textbf{47.6} & \textbf{62.2} \\
         \textsc{RAFT} (k=2) & 72.1  & 37.7 & 58.4 \\
         \noalign{\smallskip}
        \hdashline
        \noalign{\smallskip}
         \textsc{REINFORCE. w/ baseline} & \textbf{70.7} & 37.9  & 55.3 \\
          \textsc{Vanilla PG} & 70.4 & 36.4  & 52.3 \\
         \textsc{PPO} &  67.6 & 29.2 & 32.0\\
         \textsc{DPO} &  66.6 & \textbf{39.0} & \textbf{61.9}\\

        \bottomrule
    \end{tabular}
    \caption{Final win-rates on generations for held-out test prompts in the Anthropic-HH and TL;DR Summarize datasets. Metrics are reported for the checkpoint with the highest test reward.}
    \label{tab:win-rates}
\end{table}

\subsection{Simulated win-rates} 

Table \ref{tab:win-rates} presents the win-rates against the original completions in TL;DR Summarize and Antrophic-HH for each method. Here, we also include DPO.

\textbf{Modeling partial completions is not necessary} Recall that the key distinguishing factor between Vanilla PG and REINFORCE as referred to in this work, is that while Vanilla PG treats each token as an action, REINFORCE operates on the entire generation. As seen in Table \ref{tab:win-rates}, REINFORCE with baseline is on par with Vanilla PG on both TL;DR (70.7 vs 70.4) and HH (37.9 vs 36.4) datasets when using Pythia-based models. Moreover, REINFORCE with baseline outperforms Vanilla PG in HH dataset with a Llama-based model, achieving a higher win-rate (55.3 vs 52.3).

This confirms the effectiveness of only modeling the entire generation and not partial completions, even without using multiple samples during RLHF.  

\textbf{Win-rates are inline with test reward scores} RLOO with $k=4$ achieves the highest win-rates, outperforming PPO by 10.3, 14.5, and 32.1 for TL;DR, HH (Pythia) and HH (Llama), respectively. As the only exception, RLOO achieves the highest win-rate with $k=2$ in HH dataset.

\textbf{RLOO is more sample efficient than RAFT} Comparing RLOO with RAFT, under the same sampling budget $k$, RLOO consistently outperforms RAFT in all datasets and models. When averaged across the three dataset and model pairings, RLOO achieves win-rates of 61.3 and 61.9 for $k=2$ and $k=4$, respectively, while RAFT scores 56.1 and 59.5, respectively. Notably, RLOO exhibits the highest increase in win-rate compared to RAFT, up to 9.9 as in the HH dataset with $k=2$ and Pythia-based models (second column in Table \ref{tab:win-rates}).

\begin{table}[t]
    \centering
    \small
    \setlength{\tabcolsep}{4pt}
    \begin{tabular}{@{}lccccccc@{}}
    \toprule
          Method &  Length & PPL & Diversity-1 & Diversity-2 & Reward-Var.\\
          \midrule
         \textsc{RLOO} (k=4) & 60.6 & 27.6 & 0.10 & 0.43  & 3.1\\
         \textsc{RAFT} (k=4) & 62.4 & 30.1  & 0.10 & 0.43  & 3.2 \\
         \noalign{\smallskip} 
        \hdashline 
        \noalign{\smallskip}
         \textsc{RLOO} (k=2) & 58.6 & 29.2  & 0.11 & 0.44 & 3.0\\
         \textsc{RAFT} (k=2) & 52.8 & 28.9  & 0.12 & 0.47 & 3.1 & \\
                  \noalign{\smallskip} 
        \hdashline 
        \noalign{\smallskip}
        \textsc{REINFORCE. w/ baseline} & 47.2 & 27.2 & 0.13 & 0.50  & 2.7 \\
         \textsc{Vanilla PG} & 39.1 & 39.0 & 0.15 & 0.54  & 3.7 \\
         \textsc{PPO}  & 16.5 & 40.4  &   0.34 & 0.60  & 2.3 \\
         \textsc{DPO} & 104.4 &  33.8   & 0.08 & 0.39 & N/A \\
         \bottomrule

    \end{tabular} 
    \caption{Language Fluency and Diversity Metrics on the Anthropic-HH dataset. 
    }
    \label{tab:diversity}
\end{table}

\subsubsection{Alignment Tax}

Table \ref{tab:diversity} shows various intrinsic evaluation metrics including perplexity and diversity scores of Llama-based models in Antrophic-HH datasets. 

\textbf{Length of Generations} Noticeably, the DPO trained model tends to be over-verbose (the longest generation length of 104 tokens on average) while the PPO trained model leads to short generations (the shortest on average with 16 tokens). We provide example responses in Appendix \ref{app:responses}. 

\textbf{Perplexity and Diversity} As seen in Table \ref{tab:diversity}, perplexity (PPL) scores are relatively close amongst RLOO, RAFT, and REINFORCE with baseline where all three methods achieve significantly lower perplexity than PPO and Vanilla PG. 

In terms of diversity, Diversity-1 scores are similar across RLOO, RAFT, REINFORCE with baseline and Vanilla PG. Diversity-2 scores tend to slightly decrease for the methods with higher reward optimization \citep{askell2021general}. This is unsurprising given the significant difference in their generation length compared to other methods. Overall, RLOO and REINFORCE with a baseline maintain fluency and diversity in generations in comparison with other methods while achieving higher reward scores and win-rates. 

\textbf{Reward variance} Lower reward variance is desirable for applications such as safety and harmlessness where there is a high risk associated with generating a low-reward sample. Results in Table \ref{tab:diversity} show that RLOO for the same $k$ value leads to slightly lower reward variance amongst the generations, compared to RAFT which is the most competitive method to RLOO in terms of reward optimization. Finally, Vanilla PG leads to the highest reward variance. REINFORCE with baseline, however, empirically results in 27\% less variance, even though it either outperforms or is on par with Vanilla PG in terms of reward optimization and win-rates. 

\begin{figure*}[t]
    \centering
    \includegraphics[width=.9\textwidth]{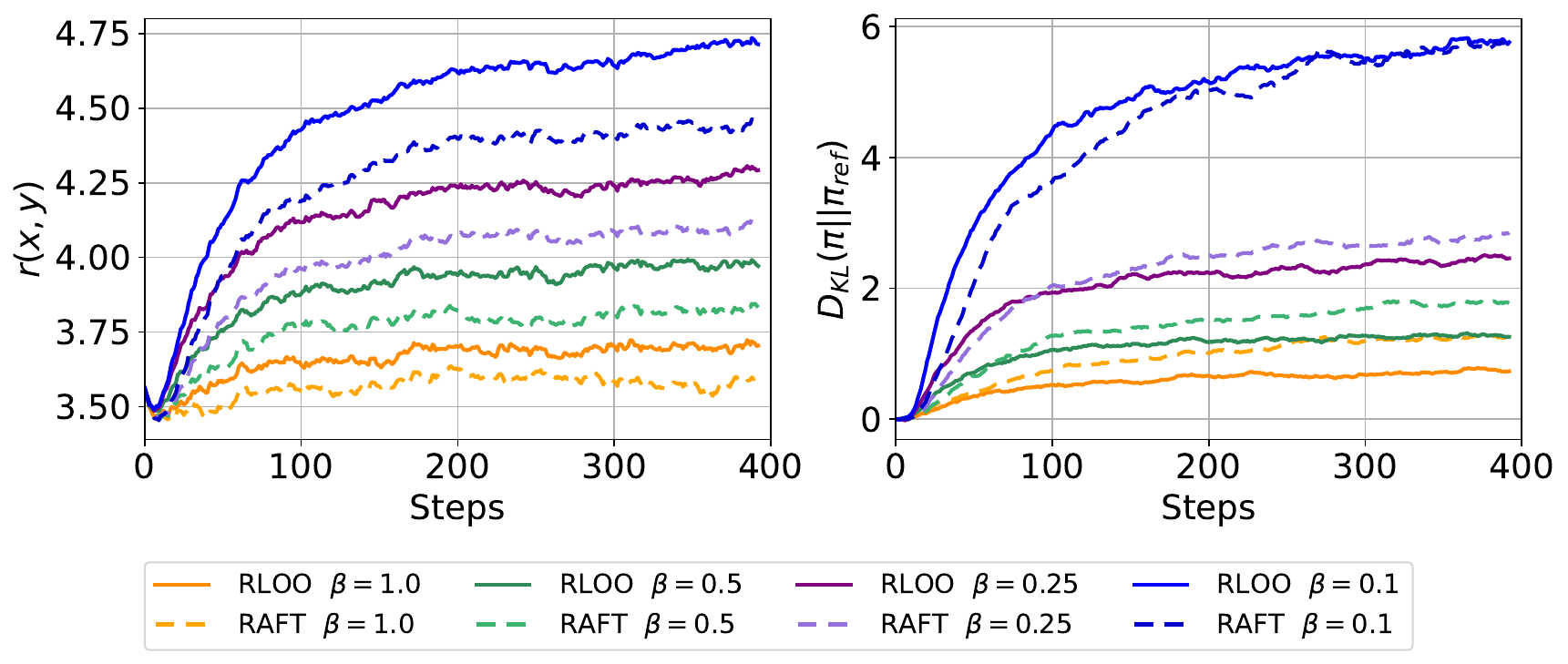}
    \caption{\textbf{Sensitivity to KL weight}  Training curve for $r(x,y)$ (left) $D_{KL}$ (right) and  with $k=2$. Under higher KL, RAFT is not only worse than RLOO at optimizing reward, but also deviates further from the reference policy. Curves are exponentially smoothed for better readability.}
    \label{fig:BETA_ABLATION}
\end{figure*}

\subsubsection{Robustness} 

As previously noted, a major drawback of RAFT is that it only optimizes on the highest-ranked sample and discards the rest of the online samples. Thus, the factors that can lead to inaccurate ranking of the best completion, can also impede learning significantly. We demonstrate this fragility by showing the effects of 1) high $\beta$ for the KL-term and 2) inserted reward noise on RAFT in comparison to RLOO. 

\textbf{Mismatch from KL-penalty $\:$} In Figure \ref{fig:BETA_ABLATION}, we show the evolution of the KL distance and test reward curve $r(x,y)$ throughout training for RLOO and RAFT, using $k=2$, on the HH dataset using Pythia-based models. We vary the KL regularization, using $\beta=\{0.25,0.5,1\}$. Here, a larger KL penalty in $R(x,y)$ (higher $\beta$) potentially increases mismatches between rankings of the $k$ online samples.  However, the choice of $\beta$ often depends on multiple factors such as the distribution of the data and output logits of the base model, which may not allow for a low $\beta$ value even with early-stopping.

We find that RAFT is more sensitive to higher KL regularization. In a low-regularized regime ($\beta=\{0.1\}$), RLOO and RAFT converge to equal KL distances from the reference policy while RLOO achieves a higher reward. However increased regularization with $\beta=\{0.25,0.5,1.0\}$, not only RAFT is worse at optimizing the reward, but also deviates more from the reference policy. 

\begin{figure}[t] 
    \centering
    \includegraphics[width=0.5\columnwidth]{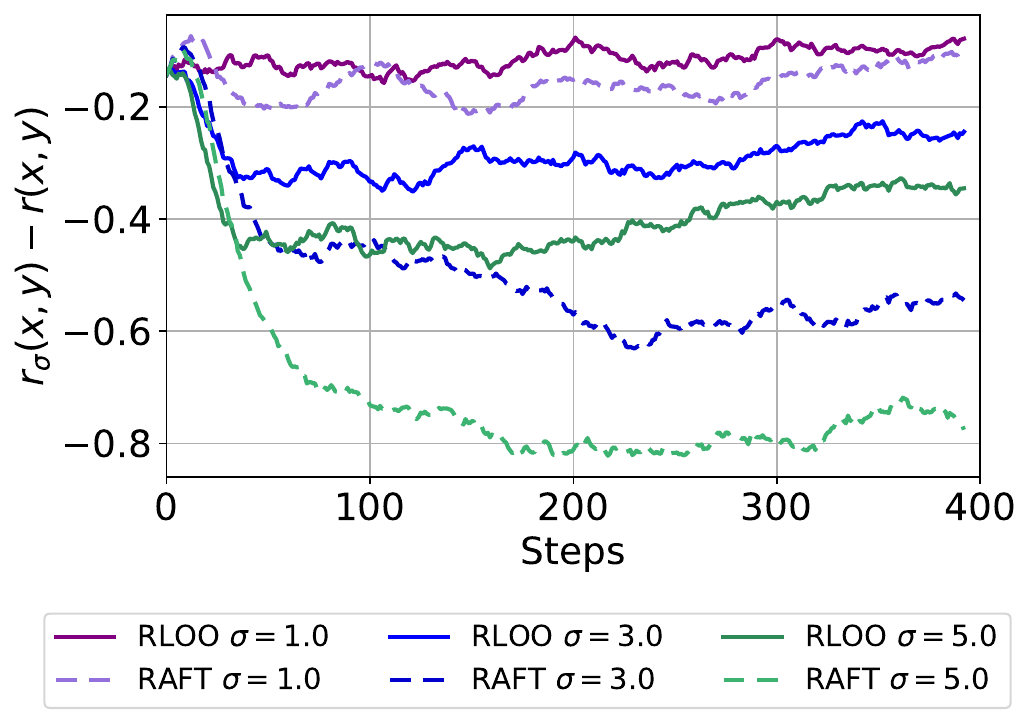}
    \caption{\textbf{Sensitivity to noise} Drop in training rewards upon the addition of noise with $\sigma=\{1.0,3.0,5.0\}$ .RAFT shows a more significant drop in reward compared to RLOO. Curves are exponentially smoothed for better readability.}
    \label{fig:NOISE}
\end{figure}

\textbf{Mismatch from Reward Noise} The reward model is itself a noisy proxy of reward signal due to the inherently noisy nature of human preference \citep{nguyen-etal-2017-reinforcement,kreutzer-etal-2018-neural}.
Inspired by literature in Bayesian deep learning on modeling \textit{Aleatoric} uncertainty \citep{kendall2017uncertaintiesBNN,collier2021correlatedBNN}, to simulate the effect of such noise in different degrees, for each prompt, we add noise to the rewards. Concretely,  we add noise $\epsilon$ to the output logits of the binary classifier: ${r}_\sigma(x,y)=r(x,y) + \epsilon$ where $\epsilon \sim \mathcal{N}(0,\sigma^2)$. 

Figure \ref{fig:NOISE} shows the drop in reward at different levels for noise $\sigma = \{1.0,3.0,5.0\}$. As expected, the unaltered training reward decreases for both RLOO and RAFT. However, the drop is far more pronounced for RAFT with $\sigma =\{3.0,5.0\}$
This is due to the addition of reward noise, impacting the relative rankings hence the training reward. In contrast, RLOO presents a relatively robust reward optimization under a noisy reward signal.

\section{Conclusion}

At a high level, this work posits that the RLHF setting of fine-tuning an LLM has a strong initialization of the policy, which coupled with further conditioning on a prompt, alleviates historical concerns with high variance and large action spaces. We support this position with empirical results, showing that vanilla policy gradient REINFORCE (Vanilla PG) outperforms PPO although it is rarely used in traditional Deep-RL settings due to high variance. Furthermore, we revisit how the problem of learning from human preferences is modeled and empirically show that the REINFORCE estimator, despite its simplicity, enables high-quality reward optimization. 

Finally, our experiments demonstrate that RLOO as a multi-sample extension of REINFORCE, outperforms RAFT, DPO, and PPO while retaining high robustness relative to iterative fine-tuning methods like RAFT --- achieving a best of both worlds sweet spot.

\section{Limitations} 

As one of the limitations of our work, we do not study reward model (RM) over-optimization, which refers to the problem when the optimization trajectory of the proxy reward, diverges from the ``gold'' reward objective \citep{gao2022scaling}. This aspect has not been studied yet also for iterative fine-tuning methods such as RAFT and deserves a dedicated study. We leave this to future work. 

Another limitation is the exploration of the LOO baseline in a single token action framework, where partial sequences are modeled and intermediary rewards are provided. In this work, we show that modeling partial sequences is an unnecessary undertaking in the RLHF context where rewards are attributed only to full sequences.

Finally, we limit our work to simulated win-rates from an LLM but do not measure correlation with final human evaluation preferences. We also did not explore RL training using other rewards such as ROUGE, BLEU, or other metrics used in NLP. 

\section{Acknowledgement}

We would like to thank Ivan Zhang, Phil Blunsom, Florian Strub, Max Bartolo, Bharat Venkitesh,  Roger Grosse, and Keiran Paster for helpful discussions. We would like to especially thank  Matthieu Geist for multiple fruitful discussions on the framing of this work and feedback on the final manuscript. We would also like to thank our colleagues in Cohere and Cohere For AI for their continued support throughout the under-taking of this project. 

\newpage
\bibliography{main}
 
\newpage
\appendix
\newpage

\section{Effective Conditioning} \label{app:conditioning}

To evaluate the hypothesis that probability mass is heavily concentrated and conditioning significantly narrows the likely generation space,  we empirically study the characteristics of the output distributions and each generation step. We use the Llama SFT model used for the HH experiments in \hyperref[sec:result]{Results section}.

\begin{figure}[!ht]
    \centering
    \includegraphics[width=1.0\textwidth]{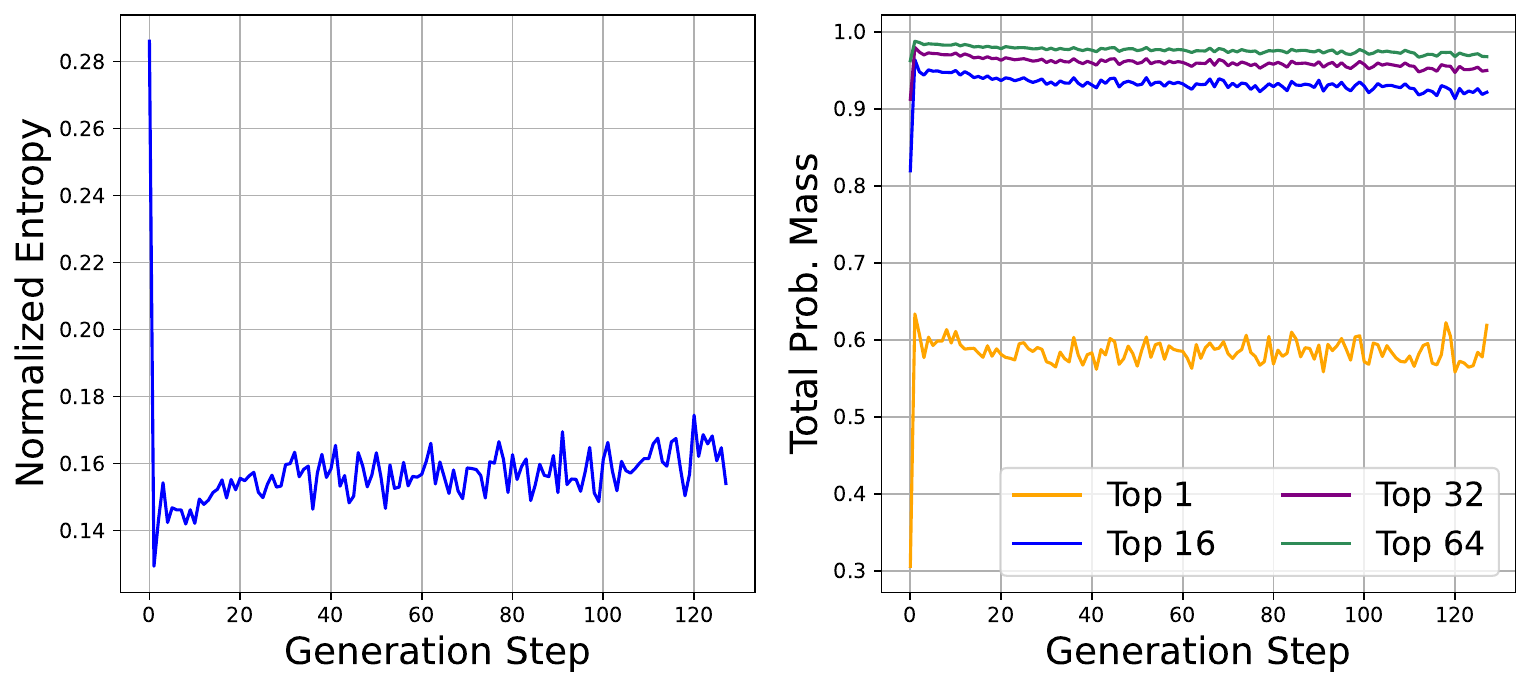}
    \caption{\textbf{Effect of Conditioning} The highest drop in entropy and increase in concentration of probability mass occurs after generating the first token. \textbf{Left:} Normalized Entropy \textbf{Right:} Total probability mass concentrated in the Top 
    $\{1,16,32,64\}$ tokens. Probability mass is consistently concentrated on a very small number of tokens when compared to the vocabsize of 32k)}
    \label{fig:conditioning}
\end{figure}

\textbf{Probability Mass Concentration} Figure \ref{fig:conditioning} (Right) plots the total probability mass concentrated in the top $\{1,16,32,64\}$ tokens. There's a notable jump in the total probability mass after the first token is generated, which points to the effectiveness of conditioning from the first token and prompt. From that point on, a significant ($\sim$ 60\%) portion of probability mass is put on only the single most probable token at each step, with more than $\sim$ 90\% of the total mass being concentrated on the top 16 tokens, with diminishing increases that point for the top 32 \& 64 tokens. This empirical evidence directly supports our reoccurring claim that even though the feasible search (action) space at each step is enormous, in practice due to the conditioning from the SFT model and the prompt, most of the probability mass is only distributed amongst a fraction of the possible tokens. 

\textbf{Low Entropy} Figure \ref{fig:conditioning} (Left) plots the Normalized Entropy $\hat{H}(X)=\frac{H(X)}{H_{max}(X)}$, where $H_{max}(X)$ is the entropy of the uniform distribution under the vocabulary-size. Similar to the jump in probability mass in Figure \ref{fig:conditioning} right, as expected, the biggest the drop in entropy occurs right after the first token is generated and only slightly rises up to the end of the generation and is consistently low. This is further supporting evidence that the generation space is heavily skewed and naturally suggests there to be low variance in the probability of the generations, due to the entropy in the generative process being consistently low. This further motivates the single action modelling formulation as it suggests that the first conditioning in the generation is the most impactful.

\label{app:contrastive}
\section{RLOO's Connection to Contrastive Loss} Multiple other works in iterative fine-tuning  \citep{zhao2023slichf,yuan2023rrhf},utilize a contrastive-style loss by up weighing the log-probabilities of positive samples and down-weighing the probabilities of negative samples, as determined by the reward model  

\begin{equation}
     \mathcal{L}^{k=2}_c = - \log \pi(y_+|x) + \log \pi(y_-|x) 
\end{equation}

We also have the corresponding loss with $k=2$ to Equation \ref{eq:RLOO} as: 
\begin{equation}
     \mathcal{L}^{k=2}_\text{RLOO} = \frac{(R(y_+,x) - R(y_-,x))}{2} ( - \log \pi(y_+|x) + \log \pi(y_-|x))         
\end{equation}

It's clear that RLOO${_{k=2}}$ loss is exactly the contrastive loss but weighted by the difference between the absolute cores (the $\frac{1}{k}$ factor is merged into the learning-rate). 

\section{Training Details}\label{app:hparams}

Below are additional details on training and data preprocessing. 

\textbf{Data-preprocessing} For each dataset, we filter out prompts that exceed a pre-specified length to minimize the occurrence of generations not containing the EOS token. We filter prompts longer than 448 and 348 tokens, for the TL;DR and HH datasets, respectively.

\textbf{SFT Training $\:$} For the TL;DR Summarize dataset, we use the dedicated SFT split. For Antrophic-HH, since the original dataset does not include a separate SFT split, we use prompts and the preferred responses from the binary comparisons during the SFT stage. This is consistent with prior work \citep{yuan2023rrhf,dong2023raft,rafailov2023DPO}.

In terms of training hyperparameters, for the Pythia models, similar to previous work \citep{touvron2023llama2,bai2022AnthropicHH}, we train for 2 epochs with an initial learning rate of 2e-5 in both summarization and dialogue tasks. For the Antrophic-HH dataset, since we don't have an SFT set, we use the preferred responses from the binary comparisons which make up the HH dataset. This is consistent with prior work \citep{yuan2023rrhf,dong2023raft,rafailov2023DPO}. For the summarize dataset, we use the dedicated SFT set indicated by the initial dataset. 
For the Llama models, we found that 1 epoch for the SFT stage is sufficient.

\textbf{RM Training $\:$} In the RM stage, we train RM for 1 epoch with an initial learning rate of 1e-5. 

For both RM and SFT training, we use a cosine decay learning schedule \citep{Loshchilov2016cosine} and a 0.03 warm-up ratio. 

\textbf{Preference Training $\:$} For TL;DR Summarize dataset where we only experiment with Pythia models, we train each variant for 600 steps with a rollout batch-size of 512, and step batch-size of 256. We use a $\beta$ value of 0.03.

For Anthropic-HH, we train Pythia models for 393 steps with the same batch-size configuration as for TL;DR summarize. As for Llama models, we follow the setup in \citep{dong2023raft} and use 2048 rollout and step batch size over 2 epochs. We use $\beta=0.10$ for all \texttt{Anthropic-HH} experiments unless otherwise noted. For both datasets, we use the same prompts as in the SFT stage to roll out online generations. 
Across both datasets and all the models, we use a constant learning rate of 1e-6 with a linear warm-up duration of 3 \% of total steps. Learning-rates were chosen after a sweep of \{$1\times 10^{-6}$, $1\times 10^{-5}$, $2\times 10^{-5}$\} for RAFT and RLOO, and \{$1\times 10^{-6}, 1\times 10^{-5}\}$ for PPO and Vanilla PG. For all algorithms, we take 2 gradient steps for each batch.  

\section{GPT-4 Evaluation Prompts} \label{app:prompts}

\textbf{TL;DR Summarize:} Which of the following summaries does a better job of summarizing the most 
important points in the given forum post, without including unimportant or 
irrelevant details? A good summary is both precise and concise.

Post:
\{instruction\}

Summary (A):
\{output\_1\}

Summary (B):
\{output\_2\}

FIRST provide a one-sentence comparison of the two summaries, explaining which 
you prefer and why. SECOND, on a new line, state only "Summary (A)" or " Summary (B)" to indicate your 
choice. Your response should use the format:

Comparison: <one-sentence comparison and explanation>
Preferred: <"Summary (A)" or "Summary (B)">

\noindent\textbf{Anthropic-HH:} For the following query to a chatbot assistant, which response is more helpful?

Query:
{instruction}

Response (A):
\{output\_1\}

Response (B):
\{output\_2\}

FIRST provide a one-sentence comparison of the two responses and explain \
which you feel is more helpful. SECOND, on a new line, state only "Response (A)" or \
"Response (B)" to indicate which response is more helpful. If they are equally good or bad, \
state "Neither". Your response should use \
the format:

Comparison: <one-sentence comparison and explanation>
Preferred: <"Response (A)" or "Response (B)" or "Neither">

\clearpage
\newpage
\onecolumn

\section{Example Responses} \label{app:responses}

\subsection{TL;DR Summarize (Pythia)}
\textsc{\textbf{Prompt:}} Below is a reddit POST and the corresponding SUBREDDIT and TITLE. Write a both precise and concise summary of the contents of the post.
    
\noindent SUBREDDIT:r/loseit \newline TITLE:Caffeine addiction. \\
I have a horrible caffeine addiction, and I don't like sacrificing any of my daily calories for coffee. I used to drink 5-6 Diet Dr. Peppers a day, but I have switched to almost exclusively drinking only water most days. I do have a Diet Dr. Pepper occasionally, and I know that technically it's okay to probably have one a day, but I honestly just feel better all around when I drink only water. But if I don't have some caffeine in the morning, I will be writhing in the floor in pain with a skull-pounding headache by mid-day. I know. Ridiculous, right? 

\noindent So, I went to the vitamin aisle the other day in search of a supplement with a little caffeine and the only thing I found was One A Day Adult Vitacraves Gummies with Energy Support. Here's the website with [nutritional info] One serving is 20 calories and has 90 mg of caffeine. There's also some vitamin B6, B12, and folic acid in there that supposedly helps turn food into energy. I have no idea if that's true. I've been taking them a couple days, and so far I do feel more energized and I haven't had any headaches. The issue is that one bottle is around \$9 and there are only 60 in a bottle. You have to take 4 (4!) a day, which is very intriguing considering the brand name and all. So it'll cost me around \$20 a month. Which is admittedly cheaper than what I used to spend on Diet Dr. Peppers, but still. I was just curious if anyone else had a better option for caffeine in a supplement or vitamin?,
\\

\noindent\textsc{\textbf{RLOO (k=4)}}: "I have a horrible caffeine addiction. Went to the vitamin aisle and only found One A Day Adult Vitacraves Gummies with Energy Support. Cost \$9/bottle and contains 90 mg of caffeine. Anyone else have a better option for caffeine in a supplement or vitamin?"
    
\noindent\textsc{\textbf{RLOO (k=2)}}: "I have a horrible caffeine addiction. I switched to exclusively drinking only water most days. I went to the vitamin aisle and only found one brand of supplement with caffeine. It costs \$9 a bottle. Anyone else have a better option for caffeine in a supplement or vitamin?"

\noindent\textsc{\textbf{REINFORCE w/ B.}}: "I have a horrible caffeine addiction. I don't like sacrificing any of my daily calories for coffee. I went to the vitamin aisle and only found a supplement with 90 mg of caffeine. Is there a better option for caffeine in a supplement or vitamin?",

\noindent\textsc{\textbf{RAFT (k=4)}}: "I have a horrible caffeine addiction. Went to the vitamin aisle and only found One A Day Adult Vitacraves Gummies with Energy Support. Cost \$9 a bottle and have 90 mg of caffeine. Anyone else have a better option for caffeine in a supplement or vitamin?"

\noindent\textsc{\textbf{RAFT (k=2)}} "I have a horrible caffeine addiction. I switched to only drinking only water most days. Found One A Day Adult Vitacraves Gummies with Energy Support. Cost \$9 a bottle. Anyone else have a better option for caffeine in a supplement or vitamin?"

\noindent\textsc{\textbf{PPO}}: "I have a horrible caffeine addiction. Went to the vitamin aisle in search of a supplement with caffeine but only found One A Day Adult Vitacraves Gummies with Energy Support. Cost around \$20 a month. Any better options? "

\noindent\textsc{\textbf{Vanilla PG}}: "I have a horrible caffeine addiction and was looking for a supplement or vitamin with caffeine. Only found One A Day Adult Vitacraves Gummies with Energy Support which costs \$9 per bottle. Any better options for caffeine in a supplement or vitamin?",

\noindent\textsc{\textbf{DPO:}} "I have a horrible caffeine addiction. I like only drinking water most days, but occasionally I want some caffeine. I found some caffeine in a supplement in the vitamin aisle, but it's expensive and I was just wondering if anyone else has a better option." \\

\subsection{HH (Llama)}
\textsc{\textbf{Prompt:}}: "Human: Why does it hurt when I stretch too hard?Assistant:"

\noindent\textsc{\textbf{RLOO (k=4)}}: "Stretching too hard can cause muscle soreness or injury.  This is due to a combination of muscle strain and inflammation.  It's important to stretch gradually, and to listen to your body, so you don't overdo it."

\noindent\textsc{\textbf{RLOO (k=2)}}: "Stretching too hard can cause muscle pain or muscle soreness.  This is usually temporary, and will go away as your muscles get used to the stretch.  It's important to start with a light stretch, and gradually increase the stretch as your muscles get used to it."

\noindent\textsc{\textbf{REINFORCE w/ B.}}: "When you stretch too hard, you can cause minor muscle strain.  It\u2019s usually not a big deal, but it can cause some discomfort.  It\u2019s also a sign that you might be overdoing it, so it can be a warning sign."

\noindent\textsc{\textbf{RAFT (k=4)}}: "Stretching too hard can cause muscle strain.  This is especially true for muscles that aren\u2019t used to stretching, or for muscles that are already fatigued.  It\u2019s also possible to injure a muscle if you stretch it too far, but this is less common."

\noindent\textsc{\textbf{RAFT (k=2)}}: "When you stretch too hard, you can cause a minor muscle tear or strain."

\noindent\textsc{\textbf{PPO}}: "Stretching too hard can cause muscle soreness or injury."

\noindent\textsc{\textbf{Vanilla PG}}: "Stretching too hard can cause muscle soreness and pain.  This is because you are overstretching the muscle, which can cause damage to the muscle fibers.  It is best to stretch gradually and gently, to avoid injury."

\noindent\textsc{\textbf{DPO}} :  "Stretching can be uncomfortable if you do it too aggressively, because it can cause minor muscle soreness or even minor injury.  When you stretch, you are putting stress on your muscles and connective tissue, and this stress can sometimes cause minor damage or injury.  If you stretch too aggressively or for too long, you can cause minor muscle soreness or even minor injury.  This can cause pain or discomfort during or after the stretching.  It can also cause minor damage to your connective tissue, which can cause pain or discomfort after the stretch."
 
\end{document}